\pdfoutput=1

\documentclass[11pt]{article}

\usepackage[]{acl}
\usepackage{times}
\usepackage{latexsym}

\usepackage[T1]{fontenc}

\usepackage[utf8]{inputenc}

\usepackage{microtype}

\usepackage{algorithm}
\usepackage{algorithmic}
\usepackage{nicefrac}  
\usepackage{amsmath}
\usepackage{booktabs}
\usepackage{bm}
\usepackage{hhline}
\usepackage{comment}
\usepackage{subfigure}
\usepackage{graphicx}
\usepackage{amssymb}  
\usepackage{verbatim}

\mathchardef\mhyphen="2D

\title{An MRC Framework for Semantic Role Labeling}

\author{
Nan Wang$^{1}$, Jiwei Li$^{2,3}$, Yuxian Meng$^{2}$\\
\bf Xiaofei Sun$^{2,3}$, Han Qiu$^{4}$, Ziyao Wang$^{5}$, Guoyin Wang$^{6}$ and Jun He$^{1}$\thanks{~~Corresponding author.} \\
$^1$Key Laboratory of Data Engineering and Knowledge Engineering of MOE, \\
School of Information, Renmin University of China\\
$^2$Shannon.AI,$^3$Zhejiang University, $^{4}$Tsinghua University \\
$^{5}$University of Leeds, $^{6}$Amazon Alexa AI \\
\{nanwang, hejun\}@ruc.edu.cn, jiwei\_li@shannonai.com
}

\begin{document}
\maketitle
\begin{abstract}
Semantic Role Labeling (SRL) aims at recognizing the predicate-argument structure of a sentence and can be decomposed into two subtasks: predicate disambiguation and argument labeling. 
Prior work deals with these two tasks independently, which ignores the semantic connection between the two tasks.
In this paper, we propose to use the machine reading comprehension (MRC) framework to bridge this gap.
We formalize predicate disambiguation as multiple-choice machine reading comprehension, where the descriptions of candidate senses of a given predicate are used as options to select the correct sense.
The chosen predicate sense is then used to determine the semantic roles for that predicate, and these semantic roles are used to construct the query for another MRC model for argument labeling.
In this way, we are able to leverage both the predicate semantics and the semantic role semantics for argument labeling.
We also propose to select a subset of all the possible semantic roles for computational efficiency.
Experiments show that the proposed framework achieves state-of-the-art or comparable results to previous work.
Code is available at \url{https://github.com/ShannonAI/MRC-SRL}.
\end{abstract}

\section{Introduction}
Semantic Role Labeling (SRL) aims at recognizing the predicate-argument structure of a sentence.  
The classic PropBank-style SRL includes two tasks: predicate disambiguation and argument labeling.
Predicate disambiguation determines the specific meaning of a predicate in a given context and argument labeling identifies the arguments of the predicate and assign them with the corresponding semantic roles, where each argument is a text span in the sentence.
PropBank defines two types of semantic roles for argument labeling: core roles and non-core roles \cite{bonial2010propbank}. Core roles are required roles that are in a close relation to the main verb in a sentence, such as {\it agent} and {\it patient}. 
There are seven core roles in PropBank: A0-A5 and AA.
Non-core roles are modifiers, such as location (LOC) and time (TMP). The specific meanings of predicates and core roles are defined in the frame files.
For example, for the sentence in Figure  \ref{fig:example}, the sense id of the predicate ``beaten'' is ``beat.02'', and its three arguments span are “The stock”, “down” and “for two days”, whose roles are respectively A1, A2, TMP.
We can get the meaning of sense label ``beat.02'' and roles A1 and A2 from the frame file.

\begin{figure}
\centering
{\linespread{0.5}\selectfont
The stock has been \underline{beaten} down for two days.\par
{\small \begin{flushleft} \hspace{1em} [\hspace{1.1em} A1\hspace{1.1em}]\hspace{5.0em}[beat.02]\hspace{0.1em}[\hspace{0.3em}A2\hspace{0.3em}]\hspace{0.3em}[\hspace{1.7em}TMP\hspace{1.7em}] \end{flushleft}}}

\vspace{4pt}
\begin{tabular}{|l|l|l|} 
\hline
sense id            & \multicolumn{2}{l|}{beat.02}             \\ 
\hline
sense                  & \multicolumn{2}{l|}{push, cause motion}  \\ 
\hline
 & A0 & causer of motion                  \\ 
\cline{2-3}
roles                       & A1 & thing moving                      \\ 
\cline{2-3}
                       & A2 & direction, destination            \\
\hline
\end{tabular}
\caption{An example of SRL. A0, A1 and A2 are {\it semantic roles} for the sense id ``beat.02''. The meanings of A0, A1 and A2 are respectively ``causer of motion'', ``thing moving'' and ``direction, destination''.}
\label{fig:example}
\end{figure}
\par

In traditional methods, predicate disambiguation and argument labeling are usually solved as two independent tasks. 
These works usually rely on feature-based methods \cite{he-etal-2018-syntax,roth-lapata-2016-neural,che2010using}) for predicate disambiguation, and use span-based  \cite{ouchi-etal-2018-span,he2018jointly,li2019dependency} or BIO-based \cite{he2017deep,strubell-etal-2018-linguistically,shi2019simple}  methods for argument labeling. These methods treat different predicate senses and argument roles as different class categories, and then solve them through classification. However, since these approaches ignore the semantic information of both predicate senses and argument roles, they are unable to establish the semantic connection between the two tasks, i.e., argument roles are defined under predicate sense via the frame files. Some works \cite{cai2018full,Conia2020BridgingTG} jointly deal with these two tasks, but still cannot establish the semantic connection. We bridge this gap with an MRC framework, and we hope that the results from predicate disambiguation will contribute to argument labeling.

\begin{figure}
\centering
\begin{tabular}{|p{0.45\textwidth}|}
\hline
\textbf{Input Sentence}   \\
The stock has been \textless{} p\textgreater{} beaten \textless{}/p\textgreater{} down for two days.     \\
\rule{0pt}{16pt}
\textbf{Multiple-Choice MRC for Predicate Disambiguation}     \\
Question: What is the sense of predicate ``beaten''?    \\
A. (Cause) pulsating motion that often makes sound     \\
B. push, cause motion  \\
C. win over some competitor   \\
Answer: B     \\
\rule{0pt}{16pt}
\textbf{Extractive MRC for Argument Labeling}      \\
Question for A0: What are the arguments with meaning "causer of motion"?  \\
Answer: No Answer                                                           \\
Question for A1: What are the arguments with meaning "thing moving"? \\
Answer: the stock \\                        
Question for A2: What are the arguments with meaning "direction, destination"?    \\
Answer: down  \\
Question for TMP: What are the time modifiers of predicate ``beaten''?  \\
Answer: for two days    \\\hline                                                
\end{tabular}
\caption{An illustration of our MRC framework for Semantic Role Labeling. The meanings of predicate senses and argument roles are used for multiple-choice and extractive MRC, respectively.}
\label{fig:illustration}
\end{figure}

For PropBank-style semantic role labeling, although the specific meanings of predicate senses and argument roles are provided in the frame files, this information is seldom used due to its huge number and lack of effective ways to utilize it. Inspired by recent success in formulating non-MRC NLP tasks as MRC tasks \cite{levy2017zero, Li2020AUM}, we propose an MRC framework for SRL, which can effectively utilize the semantic information provided by frame files. First, we transform the predicate disambiguation task into multiple-choice machine reading comprehension, where the descriptions of candidate predicate senses are used as options to select the correct sense. Then, we use the result of predicate disambiguation (i.e., the predicate sense) to determine the meaning of each core role with respect to the predicate.
Lastly, we transform argument labeling into extractive machine reading comprehension, where the description of each semantic role is used to construct the query to extract the answer span within the input sentence, which serves as the argument we want. In addition, we also propose an additional module to select a subset of all possible semantic roles to improve computational efficiency. 
We provide an example (Figure \ref{fig:example}) of the MRC framework in Figure \ref{fig:illustration}.
\par

 We conduct experiments on CoNLL2005 \cite{carreras2005introduction}, CoNLL2009 \cite{hajic-etal-2009-conll}, and CoNLL2012 \cite{pradhan2013towards} benchmarks. Experimental results show that our model can achieve SOTA or comparable results to previous works on the three benchmarks. 
 

\section{Related Work}
\subsection{Semantic Role Labeling}
Early semantic role labeling methods focused on feature engineering \cite{zhao2009multilingual,pradhan2005semantic}. Recently, neural network based models have been studied and achieved promising performance. \citet{collobert2011natural} proposed a unified neural network architecture and can avoid task-specific engineering. \citet{Zhou2015EndtoendLO} proposed to use BiLSTM as an end-to-end system for SRL. \citet{Tan2018DeepSR} applied self-attention \cite{Vaswani2017AttentionIA}  mechanism to directly draw the global dependencies of the inputs. \citet{shi2019simple} presented a BERT \cite{Devlin2019BERTPO} based model for semantic role labeling. \citet{jindal2020improved} propose a parameterized neighborhood memory adaptive method for SRL. \citet{kalyanpur2020open,tanl,Blloshmi2021GeneratingSA} cast SRL to a generative translation problem. \citet{Zhou2020ParsingAS, Marcheggiani2020GraphCO} incorporates syntactic information into SRL.
\par
Some works also show that predicate disambiguation is helpful for argument labeling. \citet{che2010improving} incorporated a word sense feature to improve the SRL performance. \citet{che2010jointly,cai2018full,Conia2020BridgingTG} jointly dealt with  predicate disambiguation and  argument labeling. These methods are different from ours and cannot use this semantic information of the sense label and role label.

\subsection{Machine Reading Comprehension}
According to the type of the answer, machine reading comprehension can be divided into the following four categories: extractive \cite{rajpurkar2016squad}, multiple-choice \cite{lai2017race}, close style \cite{onishi2016did},  and free-form \cite{nguyen2016ms}. Related to our work are extractive and multiple-choice MRC. For extractive reading comprehension such as SQuAD \cite{rajpurkar2016squad}, the answer is a span in the text, and the MRC model \cite{Seo2017BidirectionalAF} gets the answer by predicting the probability that the word is start or end. Some datasets such as DROP \cite{dua2019drop} have answers that include multiple spans, and the answers can be obtained by using BIO tagging \cite{segal2019simple}. For multiple-choice reading comprehension where the answer is one of several options, a method \cite{pan2019improving} is to calculate the score for each option and then select the option with the highest score.

\subsection{Formalizing Non-MRC Tasks as MRC}
Previous studies achieved great performance boost by applying the MRC framework to NER, dependency parsing and other non-MRC tasks.
\citet{he2015question} introduced the task of question-answer driven semantic role labeling without predefining an inventory of frames. \citet{levy2017zero} showed that relation extraction can be reduced to answering simple reading comprehension questions. \citet{McCann2018TheNL} framed ten tasks as question answering. \citet{Li2020AUM} proposed to
formulate named entity recognition as an MRC task.  Other examples include joint entity relation extraction \cite{Li2019EntityRelationEA}, coreference resolution \cite{Wu2020CorefQACR}, event extraction \cite{Li2020EventEA},  entity linking \cite{Gu2021ReadRS}, dependency parsing \cite{gan2021dependency}, text classification \cite{chai2020description}, etc. 

Our approach to formalizing argument labeling as extractive MRC is similar to QA-SRL \cite{he2015question}, but we focus on improving the performance of the model on Propbank-style SRL, while \citet{he2015question} aims to provide a new SRL annotation paradigm, and \citet{he2015question} neither uses the predicate sense definitions nor the argument role definitions provided in the frame files.

\begin{algorithm}[!ht]
	\caption{MRC framework for SRL}
	\label{alg}
	\begin{algorithmic}[1]
        \REQUIRE sentence $x=\{x_1,...,x_n\}$ with marked predicate $p$, frame files, annotation guidelines
        \ENSURE predicate sense $\hat{s}$, arguments $\bm{A}$
        \STATE Get the lemma $l$ of $p$ using SpaCy
        \STATE Get all the predicate senses $\bm{S_l}$ of $l$ and the corresponding descriptions $\bm{D^s_l}$ from the frame files
        \FOR{$s_i$ in $\bm{S_l}$}
            \STATE Get the description $d^s_i$ of sense $s_i$ from $\bm{D^s_l}$
            \STATE Concatenate $d^s_i$ and $x$ to get the input for RoBERTa
            \STATE Compute the score of $s_i$ as the answer with Eq.(\ref{eq:pd})
        \ENDFOR 
        \STATE Select the highest scoring $\hat{s}\in \bm{S_l}$ as the predicate sense of $p$
        \STATE Get the candidate argument roles $\bm{R_p}$ of $p$ from the role prediction module
        \FOR{$r_i$ in $\bm{R_p}$}
            \IF{$r_i$ is core role}
            \STATE Get the description $d^r_i$ of role $r_i$ from the frame files with $\hat{s}$
            \ELSE
            \STATE Get the description $d^r_i$ of role $r_i$ from the annotation guidelines
            \ENDIF
            \STATE Construct query $q_i$ using $d^r_i$ and $p$
            \STATE Concatenate $q_i$ and $x$ to get the input for RoBERTa
            \STATE Calculate the probability that each word in $x$ belongs to the BIO tags
        \ENDFOR
        \STATE Decode with non-overlap constraint to get the arguments $\bm{A}$ of $p$
        \RETURN $\hat{s}$, $\bm{A}$
	\end{algorithmic}
\end{algorithm}

\section{Method}
\subsection{Overview}
An overview of our model is shown in Algorithm \ref{alg}. Given a sentence $x=\{x_1,...,x_n\}$ and the predicate $p$, the predicate disambiguation task is to determine the predicate sense $s\in \bm{S}$ of $p$, where $\bm{S}$ is the set of all predicate senses, and the argument labeling task is to find all the arguments $\bm{A}=\{a_1,...,a_k\}$ of $p$, where $a_i \in \bm{A}$ is a text span in $x$, and assigning them the corresponding semantic roles.
\par
Our framework consists of three modules: predicate disambiguation, role prediction, and argument labeling, all of which use RoBERTa \cite{liu2019roberta}  as the backbone and use two special symbols
\textless{} p\textgreater{} and \textless{}/p\textgreater{} 
to mark the predicate $p$ in the input sentence $x$.
The predicate disambiguation module is intended to obtain the predicate sense of the predicate $p$. Note that we do not use the predicate sense for argument labeling directly, but only use it to get the meanings of the argument roles in the frame files.
The role prediction module is to obtain the set of candidate roles for the predicate, and the main purpose of this module is to reduce the number of questions that need to be constructed when solving the argument labeling problem via an extractive MRC.
The argument labeling module is used to obtain the arguments of the predicate, which is the core module in the whole framework.

\subsection{Multiple-Choice MRC for Predicate Disambiguation}
For the predicate disambiguation task, determining the sense label of the predicate involves two steps: identifying the lemma of the predicate, and determining the sense index of the predicate under this lemma. We use spaCy \cite{spacy} to identify the lemma of the predicate. If the recognized lemma is not in the frame files, we use the lemma with the smallest edit distance of the predicate. After identifying the lemma, we can find all the senses defined under this lemma from the frame files, and then we choose the correct sense through multiple-choice reading comprehension.
\par
Specifically, let $\bm{S_l}$ be all possible senses for the detected lemma. For each sense $s_i\in \bm{S_l}$, the corresponding sense description is $d^s_i$. We treat $d^s_i$ as option, and the input for the RoBERTa is the concatenation of $d^s_i$ and $x$. The confidence score of $s_i$ as the correct sense is calculated by:
\begin{equation}
    P(s_i=1|d^s_i, x,p)={\rm sigmoid} ({\rm FFN_p}(\bm{h^d}))  \label{eq:pd}
\end{equation}
where $\bm{h^d}$ is the context representation of the first input token from RoBERTa and ${\rm FFN_p}$ is a single layer feedforward  neural network. We train the model using the binary cross-entropy loss function. 
\footnote{We also tried to use softmax to get the probability of all senses, and then use the multi-class cross entropy loss for training, but we found the loss is unstable and hard to optimize.}
During inference, we choose the sense with the highest probability score among all the sense options as the answer.
\par

\subsection{Role Prediction}
\label{sec:role}
In semantic role labeling, most semantic roles do not have corresponding arguments given a specific input sentence. For example, in the CoNLL2005 dataset, there is a total number of 20 roles, but on average there are only 2.5 roles per predicate. Therefore, we use a role prediction module to avoid asking questions about impossible roles at the next argument labeling stage, reducing the amount of calculation required when using the MRC-based method.
\par
Let $\bm{R}$ be the set of all semantic roles (in CoNLL 2005 the size of $\bm{R}$ is 20), the purpose of role prediction is to predict a set of possible roles $\bm{R_p} \subseteq \bm{R}$ for the predicate $p$. The input to RoBERTa is the sentence $x$  with the marked predicate $p$. Let $\bm{h^r}$ be the context representation of the first token of the input sequence from RoBERTa, and $r_i\in \bm{R}$ is the i-th role of $\bm{R}$. We use the sigmoid function to calculate the probability that the predicate $p$ has a role $r_i$:
\begin{equation}
    P(r_i=1|x,p) = {\rm sigmoid}({\rm FFN_{r_i}}(\bm{h^r}))
\end{equation}
where ${\rm FFN_{r_i}}$ is a single layer feedforward  neural network. We use the binary cross entropy loss function to train the model. During inference, we only keep up to $\lambda N$
roles with the highest probability score,  where $N$ is the number of predicates in the dataset. 
\footnote{An alternative strategy is to use a fixed threshold, which performs similarly to ours. But our strategy can directly get the number of argument roles, which helps to analyze the amount of computation needed in argument labeling.}
Note that here we select the roles with the top $\lambda N$ probability scores on the whole dataset, not on the input sentence. And in the argument labeling module, we use the predicted roles from the role prediction module instead of the gold roles for training.

\subsection{Extractive MRC for Argument Labeling}
\label{sec:labeling}
We formalize argument labeling as extractive reading comprehension, where the meaning of argument role is used to construct the query, and since the answer may contain multiple spans, we use BIO tagging to extract the arguments. 
\footnote{For dependency semantic role labeling, since pre-trained language models such as BERT split a word into multiple sub-words, which is similar to span, BIO tagging is also applicable.}
In ProbBank-style SRL, a role may be a norm role, a reference role, or a continuation role. A norm role is a standard role defined in the annotation guidelines, a reference role is a reference to some other arguments, and a continuation role is a continuation phrase of a previously started argument. For example, role A1 may be N-A1 or R-A1, or C-A1. Since the subcategories of N/R/C do not contain semantic information, we do not encode such information into the query of the MRC model. We use BIO tagging to get the arguments of the predicate, and the set of BIO tags is
\begin{equation}
    \bm{T} = \{B,I\}\times\{N,R,C\}\cup\{O\}
\end{equation}
We use templates to construct the query of the MRC model. For core roles, our template is \textit{``What are the X arguments of predicate Y with meaning Z?''}, where X is the role type, Y is the predicate, and Z is the description of role X in the frame files.
For non-core roles, our template is \textit{``What are the W modifiers of predicate Y?''}, where W is the specific meanings of non-core roles defined in the annotation guidelines.
\par
Specifically, let $q_i$ represent the query corresponding to the predicate $p$ and the role $r_i \in \bm{R_p}$ , the input of the MRC model is the concatenation of $q_i$ and $x$. The context representation of $x$ in the input $<q_i, x>$ pair is $\bm{h^{r_i}}=\{\bm{h^{r_i}_1},...,\bm{h^{r_i}_n}\}$, our goal is to predict $\bm{y}^{r_i}=\{y^{r_i}_1,...,y^{r_i}_n\}$. Each $y^{r_i}_j \in y^{r_i}$ belongs to one tag in the tag set $\bm{T}$. For  $y^{r_i}_j$, its probability distribution on BIO tag set is calculated by a softmax layer:
\begin{equation}
    P(y^{r_i}_j=t|x,p,r_i) \propto {\rm exp}({\bm{W_{t}}}\bm{h^{r_i}_j}+\bm{b_t})
\end{equation}
where $t \in \bm{T}$ is a BIO tag, $\bm{W_{t}}$ and $\bm{b_{t}}$ are the corresponding parameters. We use multi-class cross entropy loss to train the model. And we use the method in section \ref{sec:decode} to get the argument.
\par
Note that at this stage, we use the predicate sense extracted at the predicate disambiguation stage to find the sense of each role selected at the role prediction stage. 
For example, suppose the predicate sense is ``beat.02'' and the semantic role is A0 as shown in Figure \ref{fig:example}, we will immediately obtain the role's sense ``causer of motion''.
In this way, the predicate sense can be leveraged for role sense detection, and thus further for semantic labeling, bridging the gap between the two tasks via a MRC framework.

\subsection{Constrained Decoding}
\label{sec:decode}
There are many global constraints in semantic role labeling \cite{punyakanok2008importance,li2020structured}, such as all arguments of the predicate cannot overlap and each core role should appear at most once for each predicate. Our MRC approach does not directly model these constraints and can not guarantee that the obtained results satisfy these constraints. For simplicity, we only consider the non-overlap arguments constraint. The previous approach of using BIO tagging \cite{he2017deep, shi2019simple} to extract arguments can naturally model the non-overlap constraint, since each word in $x$ can only belong to one of the BIO tags, there will be no overlapping words between the argument elements. But in 
our MRC-based BIO tagging method, since we have $\bm{R_p}$ roles, each word has at most $\bm{R_p}$ BIO tags.
We implement the non-overlap constraint by mapping the local role-related BIO tag of each word into a global BIO tag set. 
\par
Specifically, for the sentence $x=\{x_1,... ,x_n\}$, the goal of constraint decoding is to obtain the corresponding tag sequence $y=\{y_1,... ,y_n\}$, where $y_j \in y$ belongs to the tag set $\bm{T_p}$:
\begin{equation}
\begin{aligned}
    \bm{T_p} & = \bm{R_p} \times \{B,I\} \times \{N,R,C\} \cup \{O\} \\
\end{aligned}        
\end{equation}
For tag $t_p \in \bm{T_p}$, when it is a BI tag, it can be expressed as $r_i\mhyphen t$, where $r_i \in \bm{R_p}$ and $t \in \bm{T}$. For BI tags, we add a role tag directly before the original BI tag. For example, the B-R tag of role A1 will be converted to A1-B-R, and then the score of the new tag is equal to the probability of the original tag:
\begin{equation}
\begin{aligned}
    s(y_j=t_p)&=s(y_j=r_i\mhyphen t) \\
    &=p(y^{r_i}_j=t)
\end{aligned}   
\end{equation}
where $s(\cdot)$ is the score function. For O tags, we merge the O tags of different roles into one O tag, and the score of O tag after merging is the product of the O tag probabilities of all roles. 
\begin{equation}
 s(y_j=O)= \prod_{i=1}^{|\bm{R_p}|} p(y^{r_i}_j=O)    
\end{equation}
During inference, for each word $x_i$, its tag $y_j$ is the highest scoring tag in the new BIO tag set $\bm{T_p}$.
\begin{equation}
y_j = \mathop{\arg\max}_{t_p \in T_p} s(y_j=t_p) 
\end{equation}
And we use the BIO tag sequence $y$ to get all the arguments.
\section{Experiments}
\begin{table}
\centering
\small 
\begin{tabular}{lccc} 
\toprule
Model                    & Dev   & WSJ   & Brown  \\ 
\midrule
\citet{shi2017joint} & -     & 93.4 & 82.4  \\
\citet{roth-lapata-2016-neural}     & 94.8 & 95.5 & -      \\
\citet{he-etal-2018-syntax} & 95.0 & 95.6 & -      \\
\citet{shi2019simple}{\scriptsize+BERT} & 96.3 & 96.9 & 90.6  \\ 
\midrule
Ours{\scriptsize+BERT}   &   96.3  &  97.2  & \textbf{91.9} \\
Ours{\scriptsize+RoBERTa}  & \textbf{96.6} & \textbf{97.3} & 91.3 \\
Ours{\scriptsize -semantics} &  96.2  & 96.7 & 89.9 \\
\bottomrule
\end{tabular}
\caption{Predicate disambiguation results on CoNLL2009.}
\label{disam}
\end{table}

\begin{table}
\centering
\resizebox{0.48\textwidth}{!}{
\begin{tabular}{lcccccc}
\toprule
 & \multicolumn{3}{c}{CoNLL09 WSJ} & \multicolumn{3}{c}{CoNLL09 Brown} \\
\cmidrule(lr){2-4}\cmidrule(lr){5-7}
Model & P & R & F1 & P & R & F1 \\
\midrule
\textbf{\textit{syntax-aware}}\\
\citet{Cai2019SemiSupervisedSR} & 91.7 & 90.8 & 91.2 & 83.2 & 81.9 & 82.5 \\
\citet{Kasai2019SyntaxawareNS} &90.3 & 90.0 & 90.2 & 81.0 & 80.5 & 80.8 \\
\citet{Zhou2020ParsingAS}{\scriptsize +BERT} & 91.2 & 91.2 & 91.2 & 85.7 & 86.1 & 85.9 \\
\citet{chen-etal-2022-modeling}{\scriptsize +BERT} & 92.3 & 91.8 & 92.1 & 87.0 & 86.0 & 86.3\\
\hhline{=======}
\textbf{\textit{syntax-agnostic}} \\
\citet{li2019dependency}    & 89.6 & 91.2    & 90.4     & 81.7    & 81.4  & 81.5       \\
\citet{Conia2020BridgingTG}{\scriptsize +BERT} & 92.5 & 92.7 & 92.6 & - & - & 85.9 \\
\citet{shi2019simple}{\scriptsize +BERT}      & 92.4    & 92.3    & 92.4     & 85.7    & 85.8     & 85.7      \\
\citet{jindal2020improved}{\scriptsize +BERT}      & 90.0   & 91.5     & 90.8    & 83.5    & 86.5      & 85.0   \\ 
\midrule
\textbf{Ours}{\scriptsize +BERT} & 93.3 & 92.7 & 93.0 & 87.5 & 86.6 & 87.0 \\
\textbf{Ours}{\scriptsize +RoBERTa} &\textbf{93.5}&\textbf{93.1}&\textbf{93.3}&\textbf{87.7}&\textbf{86.6}&\textbf{87.2} \\
\bottomrule
\end{tabular}
}
\caption{Argument labeling results on CoNLL2009.}
\label{tab:dep}
\end{table}

\subsection{Datasets}
We conduct experiments on CoNLL2005 \cite{carreras2005introduction}, CoNLL2009 \cite{hajic-etal-2009-conll} and CoNLL2012 \cite{pradhan2013towards} datasets. The CoNLL2005 and CoNLL2012 datasets are span-based SRL, where the arguments are constituents (spans) in the sentence, and the CoNLL2009 dataset is dependency-based SRL, where the arguments are syntactic heads. The CoNLL2005 dataset consists of sections of the Wall Street Journal part of the Penn TreeBank, where section 2-21 is used for training, section 24 is used for development, and section 23 is used for evaluation. In addition, it also includes three sections of the Brown corpus to test the robustness of the systems. The CoNLL2009 dataset uses the same corpus as CoNLL2005, but uses NomBank to extend the annotations. The CoNLL2012 dataset is extracted from the OntoNotes v5.0 corpus. The frame files are available as official resources in the three datasets and can be used by all systems.

\subsection{Experiment Setup}
For data preprocessing  we follow \citep{li2019dependency}.
We use RoBERTa Large as the base encoder and we use two special symbols \textless{} p\textgreater{} and \textless{}/p\textgreater{}  to mark the predicate of the input sentence.
We adopt Adam as optimizer, and the warmup rate is 0.05, the initial learning rate is 1e-5, the maximum number of epochs is 20, the number of tokens in each batch is 2048.  $\lambda$ is tuned on development set to ensure that the recall of the predicted roles is higher than 99\%. All the experiments were conducted on a Tesla V100 GPU with 16GB memory. 
\par
During the parsing of the raw frame files, we found that there may be multiple sense and role definitions corresponding to one sense id, which may be caused by predicates with different part of speech or other reasons. For simplicity, we directly concatenate these different definitions, so that the final definition contains all possible cases and can be regarded as a more coarse-grained definition.
\par
Predicate disambiguation is evaluated using accuracy, and argument labeling is evaluated using micro F1. The evaluation of argument labeling in CoNLL2009 also includes the results of predicate disambiguation, where the predicate sense is treated as a special kind of argument of a virtual root node.

\subsection{Main Results}

\paragraph{Predicate Disambiguation}
We evaluate the performances of predicate disambiguation on the CoNLL2009 dataset as previous work on the CoNLL2005 and CoNLL2012 datasets did not consider predicate disambiguation. The error of lemma recognition is also included in the final results. 
In Table \ref{disam}, we report the experimental results of our method when using BERT and RoBERTa as encoders. The model using RoBERTa achieves the best results on the development set and on the in-domain test set (WSJ), and the model using BERT achieves the best results on the out-of-domain test set (Brown). The performances of BERT and RoBERTa on the development and brown test sets are opposite,  which indicates that the evaluation on the development set does not fully reflect the model's generalization ability.
\par
To investigate the impact of the sense description provided by the frame file, we also give the experimental results without using this semantic information in Table \ref{disam} (``-semantics''). In this setting, we also use RoBERTa, but the predicate sense description is replaced by the corresponding numeric label (e.g., ``02'' in ``beat.02''). 
The experimental results show that the model performs worse when this semantic information is not available, especially in the out-of-domain Brown test set, where the accuracy decreases by 1.4\%. 

\begin{table*}[!t]
\centering
\resizebox{0.95\textwidth}{!}{
\begin{tabular}{lccccccccc}
\toprule
           & \multicolumn{3}{c}{CoNLL05 WSJ}        & \multicolumn{3}{c}{CoNLL05 Brown} & \multicolumn{3}{c}{CoNLL12 Test} \\ 
\cmidrule(lr){2-4}\cmidrule(lr){5-7}\cmidrule(lr){8-10}
Model       & P         & R    & F1       & P    & R    & F1     & P     & R      & F1   \\ 
\midrule
\textbf{\textit{syntax-aware}}\\
\citet{Zhou2020ParsingAS}{\scriptsize +BERT}   & 89.0   & 88.8  & 88.9   & 81.9    & 81.0   & 81.4 & -   & -   & -  \\
\citet{Mohammadshahi2021SyntaxAwareGT}{\scriptsize +BERT}  & 89.1 & 88.7 & 88.9 & 83.9 & 82.5 & 83.2 & - & - & - \\
\citet{Xia2020SemanticRL}{\scriptsize +RoBERTa}  & 88.4     & 88.8    & 88.6     & 83.1   & 83.3   & 83.2     & -         & -    & -    \\
\citet{Marcheggiani2020GraphCO}{\scriptsize +RoBERTa} & 87.7  & 88.1   & 88.0   & 80.5   & 80.7   & 80.6  & 86.5    & 87.1      & 86.8     \\
\hhline{==========}
\textbf{\textit{syntax-agnostic}} \\
\citet{li2019dependency}    & 87.9         & 87.5        & 87.7       & 80.6      & 80.4      & 80.5      & 85.7      & 86.3      & 86.0     \\
\citet{Conia2020BridgingTG}{\scriptsize +BERT}        & -        & -          & -    & -     & -   & -     & 86.9   & 87.7  & 87.3    \\
\citet{Blloshmi2021GeneratingSA}{\scriptsize +BART}   & -     & -   & -       & -      & -     & -     & 87.8      & 86.8      & 87.3     \\
\citet{shi2019simple}{\scriptsize +BERT}   & 88.6    & 89.0    & 88.8    & 81.9     & 82.1      & 82.0      & 85.9     & 87.0      & 86.5     \\
\citet{jindal2020improved}{\scriptsize +BERT}  & 88.7  & 88.0     & 87.9   & 80.3     & 80.1    & 80.2    & 86.3    & 86.8      & 86.6    \\
\citet{tanl}{\scriptsize +T5}       & -     & -    & 89.3  & -    & -     & 82.0     & -    & -   & 87.7     \\
\citet{zhang2021semantic}{\scriptsize +RoBERTa} & 89.6 & 89.7 & 89.6 & 83.8 & 83.6 & 83.7 & 88.1 & \textbf{88.6} & \textbf{88.3} \\
\midrule
\textbf{Ours}{\scriptsize +BERT}   & 89.7     & 89.0     & 89.3     & 85.9   & 83.5   & 84.7   & 88.0   & 87.7   & 87.8   \\
\textbf{Ours}{\scriptsize +RoBERTa} &\textbf{90.4}&\textbf{89.7}&\textbf{90.0} &\textbf{86.4}&\textbf{83.8}&\textbf{85.1}&\textbf{88.6}& 87.9 & 88.3  \\ 
\bottomrule
\end{tabular}
}
\caption{Argument labeling results on CoNLL2005 and CoNLL2012.}
\label{tab:span}
\end{table*}

\paragraph{Argument Labeling}
Table \ref{tab:dep} shows the results for dependency SRL, and Table \ref{tab:span} shows the experimental results for span SRL. 
Since our method is syntax-agnostic, we first compare it with the syntax-agnostic methods. Compared with previous methods,
our improvement on the in-domain WSJ test sets of CoNLL2005 and CoNLL2009 is 0.4 and 0.7, respectively, on the out-of-domain Brown test sets is 1.4 and 1.3, respectively, and we achieve comparable results on the CoNLL2012 test set.
The out-of-domain Brown test set is used to test the robustness of the presented systems, and our method achieves greater improvement on this test set , which indicates that our approach is more robust than previous syntax-agnostic approaches because of the use of role semantics. The syntax-aware method \cite{Mohammadshahi2021SyntaxAwareGT} also performs better on the Brown test set compared to the syntax-agnostic methods \cite{shi2019simple}, a similar phenomenon to ours. However, unlike the syntax-aware approach, our approach is syntax-agnostic and utilizes the semantic information provided in the frame files rather than the syntactic information of the sentence, 
and outperforms syntax-aware methods. This observation demonstrates that leveraging semantic information in frame files provides stronger robustness than syntax-aware methods for SRL.

\section{Ablation studies}
\begin{figure}
    \centering
    \includegraphics[width=0.49\textwidth]{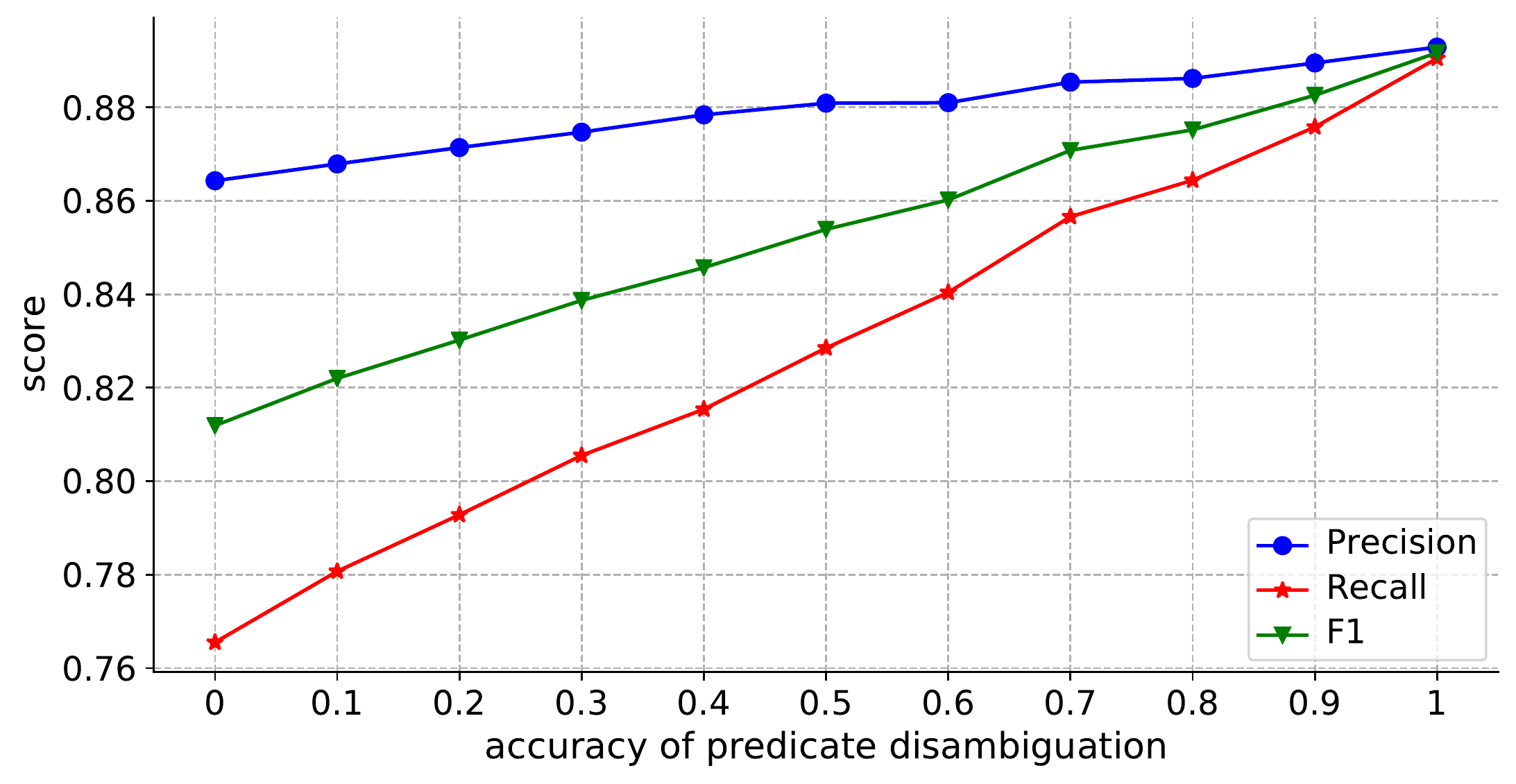}
    \caption{Experimental results on CoNLL 2005 development set with different predicate disambiguation accuracies, we use the argument labeling model trained in the main results.}
    \label{fig:sense}
\end{figure}
\subsection{Effect of Predicate Disambiguation}

Our framework uses a pipelined approach to connect the predicate disambiguation and the argument labeling task, so different predicate disambiguation accuracies may affect the results of argument labeling. Here we analyze the performance of the same argument labeling model with different predicate disambiguation accuracies. We obtain the results of different predicate disambiguation accuracies through randomly replacing part of the gold predicate senses with other predicate senses. Then we use the ordinarily trained argument labeling model to make predictions under different predicate disambiguation results. Figure \ref{fig:sense} shows that the model performs monotonically worse as the predicate disambiguation accuracy decreases, so an accurate predicate disambiguation model is required to achieve improved semantic role labeling results
\par
\subsection{Effect of Argument Role Semantics}
We also study the performance of our MRC framework in the case where the query does not contain any semantics, and in this case, the query is replaced with a category label. We use RoBERTa-Base for our experiments. When role semantics is not  considered, the F1 scores on the development set and the out-of-domain Brown test set of CoNLL2005 are 88.2 and 83.2, respectively. when role semantics is considered, the F1 scores on the development set and the out-of-domain Brown test set of CoNLL2005 are 88.5 and 83.8, respectively. The experimental results show that taking semantics into account performs better than not taking semantics into account, especially when the domains of the training and test sets are different. And this proves that the semantics of the argument roles is useful in our framework.

\subsection{Effect of Role Prediction}
\begin{table}
\centering
\begin{tabular}{lcccc} 
\toprule
Recall & 0.90 & 0.93 & 0.96 & 0.99  \\
\hline
F1     & 87.3 & 88.4 & 88.2 & 88.6  \\
\bottomrule
\end{tabular}
\caption{Experimental results of different role prediction recall scores on CoNLL2005 development set.}
\label{tab:recall}
\end{table}
Since role prediction is an upstream task of argument labeling, missing potential argument roles in the role prediction stage can lead to the error propagation problem. We mitigate this problem by ensuring that the recall of role prediction is higher than 99\% and training the argument labeling model under the predicted roles. We use RoBERTa-Base to train the model under different role recalls. Table \ref{tab:recall} shows the influence of different role prediction recall scores on argument labeling. It can be seen that when the recall is low, the F1 score of argument labeling will decrease significantly -- 87.3 when recall is 0.90 versus 88.6 when recall is 0.99.
\par

\subsection{Low-Resource Scenarios}
\begin{figure}
    \centering
    \includegraphics[width=0.48\textwidth]{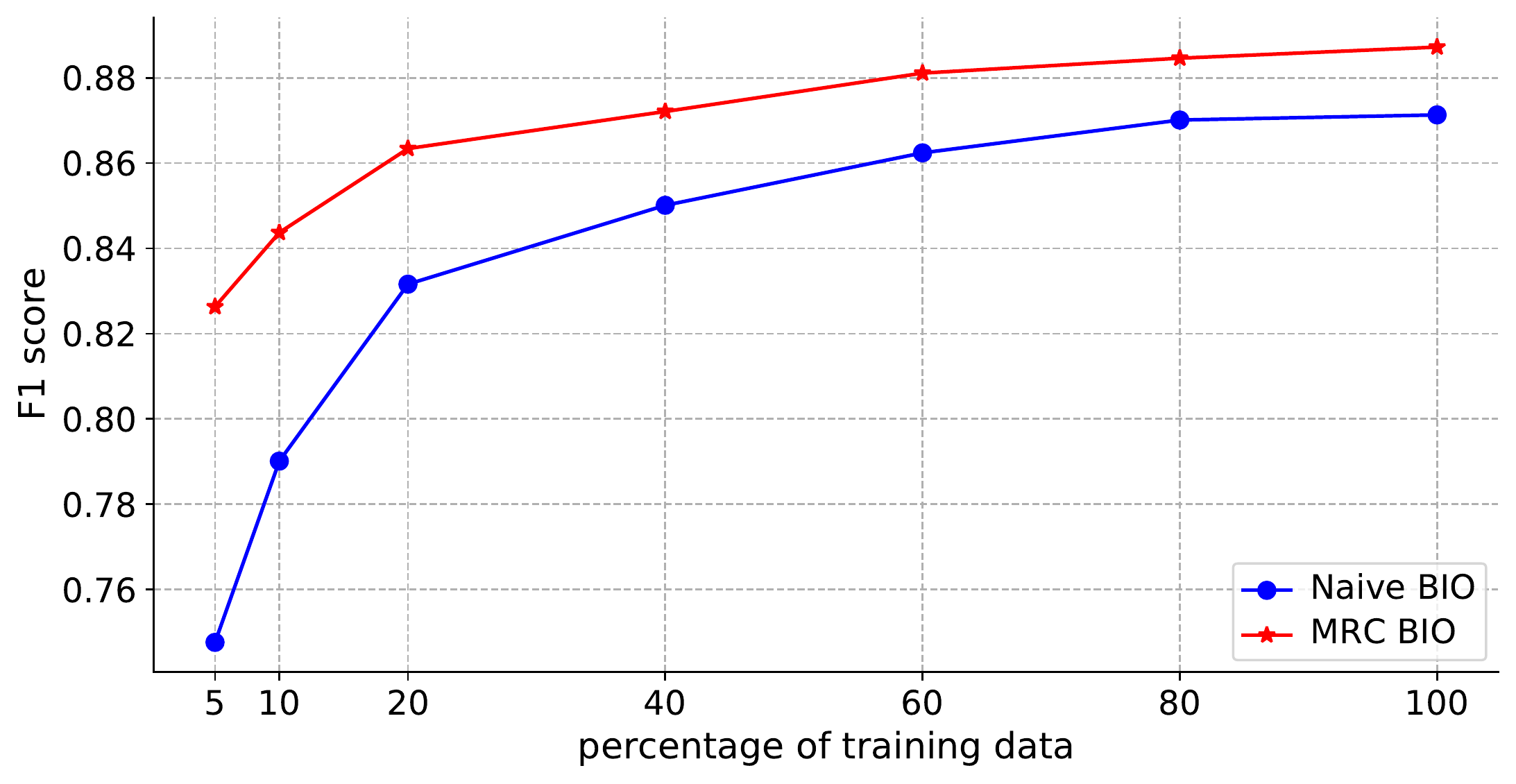}
    \caption{F1 score under different percentages of data on CoNLL2005 development set.}
    \label{fig:data}
\end{figure}
Our argument labeling model uses role semantic information, and this prior information may be helpful for model learning when the dataset is small, so we study the effect of different training data sizes. The baseline we compare is a naive BIO tagging model based on RoBERTa, since our MRC model can be seen as a simple BIO tagging model incorporating role semantic information. We use RoBERTa-Base, train on different percentages of CoNLL2005 training set, and then evaluate on CoNLL2005 development set. From Figure \ref{fig:data}, we can see that the MRC BIO model performs better in low-resource scenarios than the naive BIO model that does not use role semantics.

\subsection{Computational Overhead}
Our MRC framework performs better in robustness and low-resource scenarios due to the use of label semantic information provided in frame files, but to utilize this information, we need to encode all <label, sentence> pairs using a pre-trained model, which can be computationally intensive if the number of labels is large, we mitigate this problem by filtering impossible labels.
\footnote{We also tried to decouple the label and sentence encoding to avoid encoding the same sentence multiple times, but it did not perform as well as the simple filtering strategy.}
In predicate disambiguation, we use lemma to filter impossible predicate senses, and in argument labeling, we use an additional role prediction module to filter impossible roles. 
\par
Since the main computation in our framework is spent on the argument labeling module, here we give a rough analysis of the computational overhead it requires.
In section \ref{sec:role}, we select the $\lambda N$ roles with the highest probability scores in the dataset, which are used in the argument labeling module to construct queries, so $\lambda N$ reflects the amount of computation we need in the argument labeling module. When $\lambda=R$, this approach is equivalent to asking questions directly to all roles. In CoNLL2005, CoNLL2009, and CoNLL2012, the total number of semantic roles are 20, 20, 28, respectively, and the actual $\lambda$s in the role prediction module are 5, 4.2, 5.5, respectively, 
which indicates that our model achieves approximately 4x, 4.8x and 5.1x speedups in CoNLL2005, CoNLL2009, and CoNLL2012 compared to asking questions directly to all roles.

\section{Conclusion}
In this paper, we propose an MRC-based framework for semantic role labeling. We formalize predicate disambiguation as multiple-choice reading comprehension and argument labeling as extractive reading comprehension. Besides, we also propose a role prediction module to reduce the computation caused by considering all roles in the dataset for argument labeling.
Experiments show that our framework can effectively utilize the semantic information provided in frame files and achieve competitive results.

\section*{Acknowledgement}
We would like to thank anonymous reviewers for their comments and suggestions.
This work is supported by the Key R \& D Projects of the Ministry of Science and Technology (2020YFB2104100, 2020YFC0832500), National Natural
Science Foundation of China (62172421, 62072459).

\bibliography{custom}
\bibliographystyle{acl_natbib}

\end{document}